\documentclass[runningheads]{llncs}
\usepackage{scrextend}
\usepackage[T1]{fontenc}
\usepackage{graphicx,xcolor}
\usepackage{hyperref}
\usepackage{color}

\usepackage{graphicx,times}
\usepackage{subfig}
\usepackage{multirow}

\newcommand{\data}{\textsc{XAlignV2}}
\def\arxiv{1}

\newcommand{\mySpacing}{6pt}
\renewcommand\textfloatsep{\mySpacing}

\renewcommand\intextsep{\mySpacing}

\begin{document}
\title{\textsc{XF2T}: Cross-lingual Fact-to-Text Generation for Low-Resource Languages}

\if\arxiv1
\author{Shivprasad Sagare, Tushar Abhishek, Bhavyajeet Singh, Anubhav Sharma,\\ Manish Gupta, Vasudeva Varma\\
\texttt{\{shivprasad.sagare,tushar.abhishek\}@research.iiit.ac.in}\\
\texttt{\{bhavyajeet.singh,anubhav.sharma\}@research.iiit.ac.in\\ \{manish.gupta,vv\}@iiit.ac.in}
}
\authorrunning{Sagare et al.}
\institute{Information Retrieval and Extraction Lab, IIIT Hyderabad, India}
\else
\author{}
\institute{}
\fi

%
%

%

\maketitle              
\begin{abstract}
Multiple business scenarios require an automated generation of descriptive human-readable text from structured input data. Hence, fact-to-text generation systems have been developed for various downstream tasks like generating soccer reports, weather and financial reports, medical reports, person biographies, etc.
Unfortunately, previous work on fact-to-text (F2T) generation has focused primarily on \emph{English} mainly due to the high availability of relevant datasets. Only recently, the problem of cross-lingual fact-to-text (XF2T) was proposed for generation across multiple languages alongwith a dataset, \textsc{XAlign} for eight languages. However, there has been no rigorous work on the actual XF2T generation problem.
We extend \textsc{XAlign} dataset with annotated data for four more languages: Punjabi, Malayalam, Assamese and Oriya. We conduct an extensive study using popular Transformer-based text generation models on our extended multi-lingual dataset, which we call \data{}. Further, we investigate the performance of different text generation strategies: multiple variations of pretraining, fact-aware embeddings and structure-aware input encoding. Our extensive experiments show that a multi-lingual mT5 model which uses fact-aware embeddings with structure-aware input encoding leads to best results on average across the twelve languages. 
We make our code, dataset and model publicly available\footnote{\url{https://tinyurl.com/CrossF2TV2}\label{datafootnote}}, and hope that this will help advance further research in this critical area.

\end{abstract}

\section{Introduction}
Fact-to-text (F2T) generation~\cite{reiter1997building} is the task of transforming structured data (like fact triples) into natural language. F2T systems are vital in many downstream applications like automated dialog systems~\cite{wen2016multi}, domain-specific chatbots~\cite{novikova2017e2e}, open domain question answering~\cite{chen2020kgpt}, authoring sports reports~\cite{chen2008learning}, financial reports~\cite{plachouras2016interacting}, news reports~\cite{leppanen2017data}, etc. But most of such F2T systems are English only and not available for low-resource (LR) languages. This is mainly due to lack of training data in LR languages. 
Table~\ref{tab:wikiDataWikipediaStats} shows that structured Wikidata entries for person entities in LR languages are minuscule in number compared to that in English. Also, average facts per entity in LR languages are much smaller than in English. Thus, monolingual F2T for LR languages suffers from data sparsity. Hence,
the problem of \emph{cross-lingual F2T generation (XF2T)} is important, wherein the input is a set of English facts and output is a  sentence capturing the fact-semantics in the specified LR language~\cite{abhishek2022xalign}. Note that a fact is a triple composed of subject, relation and object.

\begin{figure}
    \centering
    \includegraphics[width=\columnwidth]{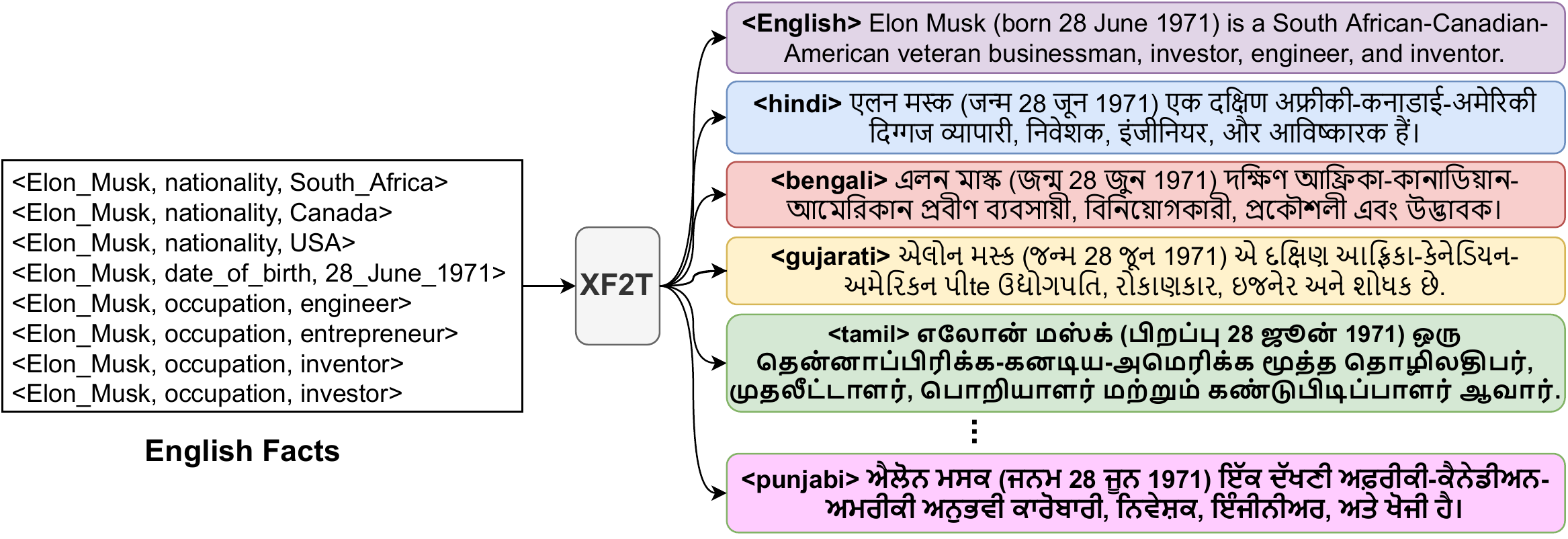}
    \caption{XF2T example: Generating English, Hindi, Bengali, Gujarati, Tamil and Punjabi sentences to capture semantics from English facts.}
    \label{fig:example}
\end{figure}

One major challenge in training an XF2T system is the availability of aligned data where English facts are  well-aligned with semantically equivalent LR text. The manual creation of such a high-quality aligned dataset requires human annotations and is quite challenging to scale.
Abhishek et al.~\cite{abhishek2022xalign} proposed transfer learning and distance supervision based methods for cross-lingual alignment. Further, they use these alignment methods to contribute the \textsc{XAlign} dataset which consists of sentences from LR language Wikipedia mapped to English fact triples from Wikidata. It contains data for the following eight languages: Hindi (hi), Telugu (te), Bengali (bn), Gujarati (gu), Marathi (mr), Kannada (kn), Tamil (ta) and English (en). In this work, we extend this dataset to four more LR languages: Punjabi (pa), Malayalam (ml), Assamese (as) and Oriya (or). Fig.~\ref{fig:example} shows an XF2T example from our extended dataset, \data{}. 

\setlength{\tabcolsep}{2pt}
\begin{table}[!t]
    \centering
    \scriptsize
    \begin{tabular}{|l|c|c|c|c|c|c|c|c|c|c|c|c|}
\hline
Lang.&hi&mr&te&ta&gu&bn&kn&pa&as&or&ml&\textbf{en}\\
\hline
\hline
|E|&26.0K&16.5K&12.4K&26.0K&3.5K&36.2K&7.5K&10.8K&1.7K&7.4K&19.9K&\textbf{79.3K}\\
\hline
|F|&271.2K&174.2K&142.5K&280.0K&38.1K&502.1K&83.3K&107.5K&2.1K&87.1K&243.2K&\textbf{1.65M}\\
\hline
F/E&10.43&10.56&11.49&10.77&10.88&13.87&11.10&9.95&1.21&11.77&12.22&\textbf{20.80}\\
\hline
|A|&22.9K&15.9K&7.8K&25.6K&1.9K&29.0K&4.5K&10.5K&1.7K&3.2K&16.7K&\textbf{72.7K}\\
\hline
    \end{tabular}
    \caption{WikiData+Wikipedia statistics for the person entities across languages. |E|=\#WikiData entities, |F|=\#Facts, F/E=Avg facts per entity, |A|=\#Wikipedia articles. Stats are only for entities in our \data{} dataset.}
    \label{tab:wikiDataWikipediaStats}
\end{table}

In their short paper, Abhishek et al.~\cite{abhishek2022xalign} focused on dataset creation and hardly on the XF2T task. In this paper, we rigorously investigate the XF2T problem. 
First, we experiment with standard existing Transformer-based multi-lingual encoder-decoder models like the vanilla Transformer, IndicBART and mT5. Next, we explore performance across various training setups: bi-lingual, translate-output, translate-input and multi-lingual. Further, we systematically explore various strategies for improving XF2T generation like multi-lingual data-to-text pre-training, fact-aware embeddings, and structure-aware encoding. Overall, we make the following contributions in this work. 
\begin{itemize}
\itemsep=0pt
\topsep=0pt
    \item We extend the \textsc{XAlign} dataset with annotated XF2T data corresponding to four more LR languages, leading to a new dataset, \data{}.
    \item We rigorously experiment with multiple encoder-decoder models, training setups, pretraining methods, and input representations toward building a robust XF2T system.
    \item We find that a multi-lingual mT5 model which uses fact-aware embeddings along with structure-aware input encoding leads to best results. Our best small-scale model achieves an average BLEU of 29.27, METEOR of 53.64, and chrF++ of 57.30 for XF2T across 12 languages. We make the code, dataset and our model publicly available\footref{datafootnote}.
\end{itemize}

\if\arxiv1
The remainder of the paper is organized as follows. We discuss related work in the areas of cross-lingual Natural Language Generation (NLG) and F2T Generation in Section~\ref{sec:relatedWork}. We discuss data collection,  pre-processing details, and basic dataset analysis for our \data{} dataset in Section~\ref{sec:dataCreation}. In Section~\ref{sec:approach}, we discuss various Transformer-based methods, different training setups and input representations. We discuss implementation details and metrics in Section~\ref{sec:experiments}. Section~\ref{sec:results} does a detailed analysis of the results. Finally, we conclude with a brief summary in Section~\ref{sec:conclusion}.
\fi
\section{Related Work}
\label{sec:relatedWork}
\if\arxiv1
In this section, we discuss related work in the areas of cross-lingual NLG and F2T Generation.
\fi
\noindent\underline{\textbf{Cross-lingual NLG}}: 
Recently there has been a lot of work on cross-lingual NLG tasks like machine translation~\cite{chi2021mt6,liu2020multilingual}, question generation~\cite{chi2020cross}, news title generation~\cite{liang2020xglue}, and  summarization~\cite{zhu2019ncls} thanks to models like XNLG~\cite{chi2020cross}, mBART~\cite{liu2020multilingual}, mT5~\cite{xue2021mt5}, etc. Abhishek et al.~\cite{abhishek2022xalign} proposed a specific cross-lingual NLG task: XF2T. In this work, we investigate effectivess of multiple modeling techniques for the XF2T task. Further, from a knowledge graph (KG) and text linking perspective, our work is related to tasks like entity linking (link mention in a sentence to a KG entity)~\cite{botha2020entity} and fact linking (linking sentence to a set of facts)~\cite{kolluru2021multilingual}. As against this, XF2T is the problem of generating a sentence given a set of facts.


\begin{table*}[!b]
    \centering
    \scriptsize
    \begin{tabular}{|l|l|c|c|c|c|c|c|}
    \hline
Dataset&Languages&A/M&$|$I$|$&F/I&$|$P$|$&$|$T$|$&X-Lingual\\
\hline
\hline
WikiBio&en&A&728K&19.70&1740&26.1&No\\
\hline
E2E&en&M&50K&5.43&945&20.1&No\\
\hline
WebNLG 2017&en&M&25K&2.95&373&22.7&No\\
\hline
fr-de Bio&fr, de&A&170K, 50K&8.60, 12.6&1331, 1267&29.5, 26.4&No\\
\hline
TREX&en&A&6.4M&1.77&642&79.8&No\\
\hline
WebNLG 2020&en, ru&M&40K, 17K&2.68, 2.55&372, 226&23.7 &Yes\\
\hline
KELM&en&A&8M&2.02&663&21.2&No\\
\hline
WITA&en&A&55K&3.00&640&18.8&No\\
\hline
WikiTableT&en&A&1.5M&51.90&3K&115.9&No\\
\hline
GenWiki&en&A&1.3M&1.95&290&21.5&No\\
\hline
\textsc{XAlign}&en + 7 LR&A&0.45M&2.02&367&19.8&Yes\\
\hline
\hline
\data{}&en + 11 LR&A&0.55M&1.98&374&19.7&Yes\\
\hline
    \end{tabular}
    \caption{Statistics of popular Fact-to-Text datasets: WikiBio~\cite{lebret2016wikibio}, E2E~\cite{novikova2017e2e}, WebNLG 2017~\cite{gardent2017webnlg}, WebNLG 2020~\cite{ferreira20202020}, fr-de Bio~\cite{nema2018generating}, KELM~\cite{agarwal2021knowledge}, WITA~\cite{fu2020partially}, WikiTableT~\cite{chen2021wikitablet}, GenWiki~\cite{jin2020genwiki}, TREX~\cite{elsahar2018trex},  XAlign~\cite{abhishek2022xalign}, and \data{} (ours). Alignment method could be A (automatic) or M (manual). $|$I$|$=number of instances. F/I=number of facts per instance. $|$P$|$=number of unique relations. $|$T$|$=average number of tokens per instance.}
    \label{tab:dataStatsSurvey}
\end{table*}

\noindent\underline{\textbf{F2T Datasets}}: Table~\ref{tab:dataStatsSurvey} shows basic statistics of popular F2T datasets. There exists a large body of work on generic structured data to text, but here we list only F2T datasets. These datasets contain text from various domains like person, sports, restaurant, airport, politician, artist, etc. Also, these datasets vary widely in terms of statistics like number of instances, number of facts per instance, number of unique relations and average number of tokens per instance. Unlike other datasets which are mostly on English only, \textsc{XAlign} and our dataset, \data{}, contain 8 and 12 languages respectively, and are cross-lingual datasets. 


\noindent\underline{\textbf{F2T Generation}}:
Training F2T models requires aligned data with adequate content overlap. Some previous studies like WebNLG~\cite{gardent2017webnlg} collected aligned data by crowdsourcing while others have performed automatic alignment by heuristics like TF-IDF. Abhishek et al.~\cite{abhishek2022xalign} explore two different unsupervised methods to perform cross-lingual alignment. We leverage the ``transfer learning from Natural Language Inference task'' based method for our work.

Initial F2T methods were template-based and were therefore proposed on domain-specific data like medical~\cite{bontcheva2004automatic}, cooking~\cite{cimiano2013exploiting}, person~\cite{duma2013generating}, etc. They align entities in RDF triples with entities mentioned in sentences, extract templates from the aligned sentences, and use templates to generate sentences given facts for new entities. Template-based methods are brittle and do not generalize well.

Recently, Seq-2-seq neural methods~\cite{lebret2016wikibio,mei2016talk} have become popular for F2T. These include vanilla LSTMs~\cite{vougiouklis2018neural}, LSTM encoder-decoder model with copy mechanism~\cite{shahidi2020two}, LSTMs with hierarchical attentive encoder~\cite{nema2018generating}, pretrained Transformer based models~\cite{ribeiro2021investigating} like BART~\cite{lewis2020bart} and T5~\cite{raffel2020exploring}.
Richer encoding of the input triples has also been investigated using a combination of graph convolutional networks and Transformers~\cite{zhao2020bridging}, triple hierarchical attention networks~\cite{chen2020kgpt}, or Transformer networks with special fact-aware input embeddings~\cite{chen2020kgpt}. Some recent work also explores specific F2T settings like plan generation when the order of occurrence of facts in text is available~\cite{zhao2020bridging} or partially aligned F2T when the text covers more facts than those mentioned in the input~\cite{fu2020partially}. However, all of these methods focus on English fact to text only. Only recently, the XF2T problem was proposed in~\cite{abhishek2022xalign} but their focus is on problem formulation and dataset contribution. In this paper, we extensively evaluate multiple methods for the XF2T generation task. 

\section{\data{}: Data Collection, Pre-processing and Alignment}
\label{sec:dataCreation}


\noindent\underline{\textbf{Data Collection and Pre-processing}}: 
We start by gathering a list of 95018 person entities from Wikidata each of which have a link to a corresponding Wikipedia page in at least one of our 11 LR languages. This leads to a dataset $D$ where every instance $d_i$ is a tuple $\langle$entityID, English Wikidata facts, LR language, LR-language Wikipedia URL for the entityID$\rangle$.
We extract facts from the 20201221 WikiData dump for all the 12 languages for each entity in $D$ using the WikiData API\footnote{\url{https://query.wikidata.org/}}. We gathered facts corresponding to only these Wikidata property (or relation) types which capture most of the useful factual information for person entities: WikibaseItem, Time, Quantity, Monolingualtext. We retain any additional supporting information associated with the fact triple as a fact qualifier. This leads to overall $\sim$0.55M data instances across all the 12 languages. 

For each language, we used the Wikiextractor~\cite{Wikiextractor2015} to extract text from the 20210520 Wikipedia xml dump. We split this text into sentences using IndicNLP~\cite{kakwani2020indicnlpsuite}, with a few additional heuristics to account for Indic punctuation characters, sentence delimiters and non-breaking prefixes. We prune out (1) other language sentences using Polyglot language detector\footnote{\url{https://polyglot.readthedocs.io/en/latest/Detection.html}}, (2) sentences with <5 words or >100 words, (3) sentences which could potentially have no factual information (sentences with no noun or verb\footnote{For POS tagging, we used Stanza~\cite{qi2020stanza} for en, hi, mr, te, ta; LDC Bengali POS Tagger~\cite{bali2010indian} for bn and ma; and~\cite{patel2008part} for gu. For other languages, we obtained a list of all entities from language-specific Wikidata, and then labeled mentions of these entities as proper nouns; for these languages we ignored the presence of verbs check.}). For each sentence per Wikipedia URL, we also store the section information.

\noindent\underline{\textbf{Fact-to-Text Alignment}}:
For every (entity $e$, language $l$) pair, the pre-processed dataset has a set $F_{el}$ of English Wikidata facts and a set of Wikipedia sentences $S_{el}$ in that language. Next, we build an automatic aligner to associate a sentence in $S_{el}$ with a subset of facts from $F_{el}$. This is done using a two-stage system, similar to~\cite{abhishek2022xalign}. The first stage (Candidate Generation) generates (facts, sentence) candidates based on automated translation and syntactic+semantic match. For every entity and language pair, Stage 1 outputs sentences each associated with a maximum of $K$ facts. We use transfer learning from NLI (Natural language Inference) task to retain only strongly aligned (fact, sentence) pairs in Stage 2. The input is ``sentence$\langle$SEP$\rangle$subject|relation|object''. Given a premise and a hypothesis, NLI aims to predict whether the hypothesis entails, contradicts or is neutral to the premise. Fact to sentence alignment is semantically similar to NLI where the sentence and the fact can be considered as the premise and the hypothesis resp. Hence, we could infer (fact, sentence) alignment by directly probing an NLI mT5~\cite{xue2021mt5} model. We use the Xtreme-XNLI finetuned mT5 checkpoint from Huggingface. If the model predicts entailment, we consider the (fact, sentence) pair to be aligned, else not. Thus, from amongst $K$ candidate facts for every sentence, we select a subset of facts. 

\begin{table}[!t]
    \centering
    \scriptsize
    \begin{tabular}{|l|c|c|c|c|c|c|c|c|c|c|}
    \hline
    &\multirow{2}{*}{|V|}&\multicolumn{3}{c|}{Train+Validation}&\multicolumn{5}{c|}{Manually Labeled Test}\\
    \cline{3-10}
& &|I|&|T|& |F|& $\kappa$&|A| &|I|&|T|&|F|\\
    \hline
    \hline
    hi&75K&57K&25.3/5/99&2.0&0.81&4&842&11.1/5/24&2.1\\
    \hline
    mr&50K&19K&20.4/5/94&2.2&0.61&4&736&12.7/6/40&2.1\\
    \hline
    te&61K&24K&15.6/5/97&1.7&0.56&2&734&9.7/5/30&2.2\\
    \hline
    ta&121K&57K&16.7/5/97&1.8&0.76&2&656&9.5/5/24&1.9\\
    \hline
    en&104K&133K&20.2/4/86&2.2&0.74&4&470&17.5/8/61&2.7\\
    \hline
    gu&35K&9K&23.4/5/99&1.8&0.50&3&530&12.7/6/31&2.1\\
    \hline
    bn&131K&121K&19.3/5/99&2.0&0.64&2&792&8.7/5/24&1.6\\
    \hline
    kn&88K&25K&19.3/5/99&1.9&0.54&4&642&10.4/6/45&2.2\\
    \hline
    pa&59K&30K&32.1/5/99&2.1&0.54&3&529&13.4/5/45&2.4\\
    \hline
    as&27K&9K&19.23/5/99&1.6&-&1&637&16.22/5/72&2.2\\
    \hline
    or&28K&14K&16.88/5/99&1.7&-&2&242&13.45/7/30&2.6\\
    \hline
    ml&146K&55K&15.7/5/98&1.9&0.52&2&615&9.2/6/24&1.8\\
    \hline
    \end{tabular}
    \caption{Basic Statistics of \data{}. |I|=\# instances, |T|=avg/min/max word count, |F|=avg \#facts, |V|=Vocab. size, $\kappa$=Kappa score, |A|=\#annotators. For Train+Validation, min and max fact count is 1 and 10 resp across languages.\protect\footnotemark}
    \label{tab:xalignDataStats}
\end{table}

\footnotetext{For or, $\kappa$ is not reported since we did not get redundant judgments done due to lack of available annotators. For as, $\kappa$ is not reported since we had only one annotator.}

\noindent\underline{\textbf{Manual Annotations for Ground-Truth Data}}:
We need manually annotated data for evaluation of our XF2T generation. We perform manual annotation in two phases. For both the phases, the annotators were presented with (LR language sentence $s$, $K$ English facts) output by Stage 1. They were asked to mark facts present in $s$. There were also specific guidelines to ignore redundant facts, handle abbreviations, etc. More detailed annotation guidelines are mentioned here\footref{datafootnote}.
In the first phase, we got 60 instances labeled per language by a set of 12 expert annotators (trusted graduate students who understood the task very well). In phase 2, we selected 8 annotators per language from the National Register of Translators\footnote{\url{https://www.ntm.org.in/languages/english/nrtdb.aspx}}. We tested these annotators using phase 1 data as golden control set, and  shortlisted up to 4 annotators per language who scored highest (on Kappa score with golden annotations) for final annotations. We report details of this test part of our \textsc{XAlignV2} dataset in Table~\ref{tab:xalignDataStats}. On average, a sentence can be verbalized using around two fact triples.

\begin{figure}[!t]
\begin{minipage}{0.48\columnwidth}
    \centering
    \includegraphics[width=\columnwidth]{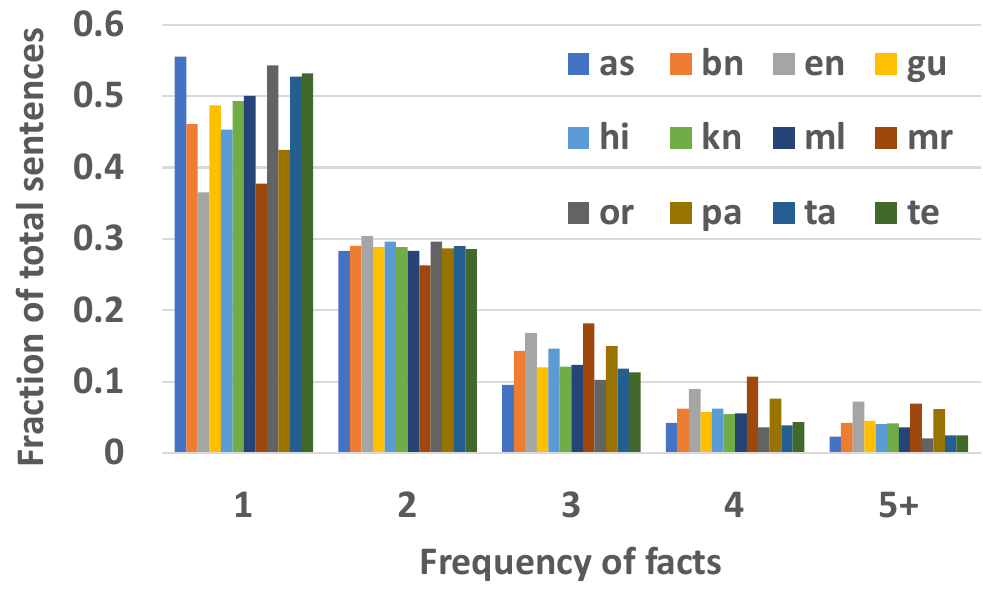}
    \caption{Fact Count Distribution across languages}
    \label{fig:language_fact_number_fraction}
    \end{minipage}
    \hspace{0.05\columnwidth}
    \begin{minipage}{0.46\columnwidth}
    \centering
    \includegraphics[width=\columnwidth]{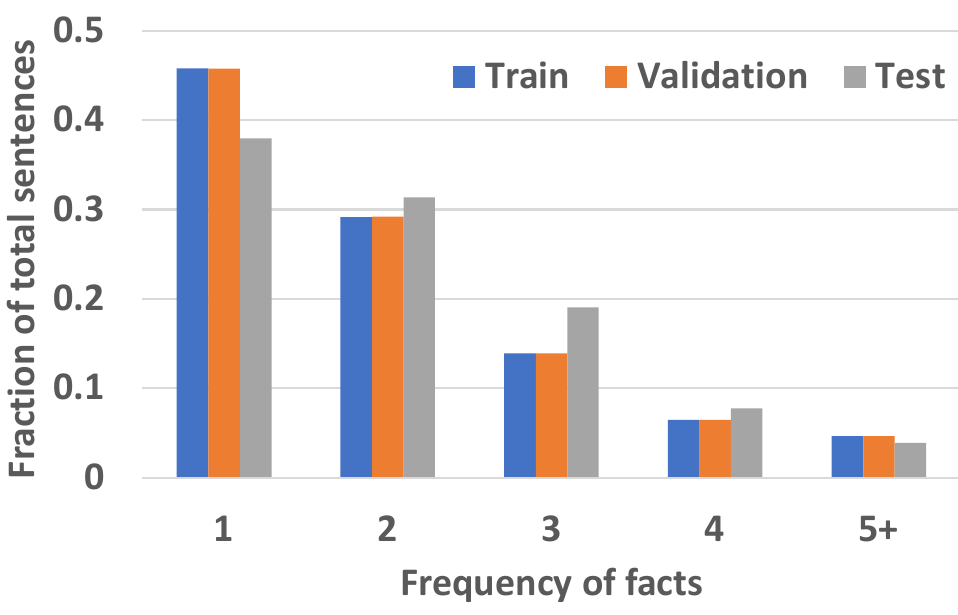}
    \caption{Fact Count Distribution across data subsets}
    \label{fig:train_test_val_factcount}
    \end{minipage}
\end{figure}

\noindent\underline{\textbf{\data{} Dataset Analysis}}: 
Using the mT5-transfer-learning Stage 2 aligner on Stage 1 output leads to the Train+Validation part of \data{}. Table~\ref{tab:xalignDataStats} shows the dataset statistics. Figs.~\ref{fig:language_fact_number_fraction} and~\ref{fig:train_test_val_factcount}  show fact count distribution. We observe that a large percent of sentences contain more than one fact across languages. Also, the distribution is similar across languages and data subsets.
Finally, Table~\ref{tab:topkpredicates} shows top 10 frequent fact relations across all the languages. 


\begin{table*}
    \centering
    \scriptsize
    \begin{tabular}{|l|p{0.95\textwidth}|}
    \hline
hi&occupation, date of birth, position held, cast member, country of citizenship, award received, place of birth, date of death, educated at, languages spoken written or signed\\
\hline
mr&occupation, date of birth, position held, date of death, country of citizenship, place of birth, member of sports team, member of political party, cast member, award received\\
\hline
te&occupation, date of birth, position held, cast member, date of death, place of birth, award received, member of political party, country of citizenship, educated at\\
\hline
ta&occupation, position held, date of birth, cast member, country of citizenship, educated at, place of birth, date of death, award received, member of political party\\
\hline
en&occupation, date of birth, position held, country of citizenship, educated at, date of death, award received, place of birth, member of sports team, member of political party\\
\hline
gu&occupation, date of birth, cast member, position held, award received, date of death, languages spoken written or signed, place of birth, author, country of citizenship\\
\hline
bn&occupation, date of birth, country of citizenship, cast member, member of sports team, date of death, educated at, place of birth, position held, award received\\
\hline
kn&occupation, cast member, date of birth, award received, position held, date of death, performer, place of birth, author, educated at\\
\hline
pa&occupation, date of birth, place of birth, date of death, cast member, country of citizenship, educated at, award received, languages spoken, written or signed, position held\\
\hline
as&occupation, date of birth, cast member, position held, date of death, place of birth, country of citizenship, educated at, award received, member of political party\\
\hline
or&occupation, date of birth, position held, cast member, member of political party, place of birth, date of death, award received, languages spoken, written or signed, educated at\\
\hline
ml&occupation, cast member, position held, date of birth, educated at, award received, date of death, place of birth, author, employer\\
\hline
    \end{tabular}
    \caption{Top-10 frequent fact relations across languages.}
    \label{tab:topkpredicates}
\end{table*}

\section{Approaches for Cross-lingual Fact to Text}
\label{sec:approach}
In this section, we first discuss our input representation. Next, we discuss various Transformer-based methods, different training setups, multiple pretraining methods, and discussion on fact-aware embeddings.

\noindent\underline{\textbf{Structure-aware Input encoding}}: 
Each input instance consists of multiple facts $F=\{f_1, f_2, \ldots, f_n\}$ and a section title $t$. A fact $f_i$ is a tuple composed of subject $s_i$, relation $r_i$, object $o_i$ and $m$ qualifiers $Q={q_1, q_2,\ldots, q_m}$. Each qualifier provides more information about the fact. Each of the qualifiers $\{q_j\}_{j=1}^m$ can be linked to the fact using a fact-level property which we call as qualifier relation $qr_j$. For example, consider the sentence: ``Narendra Modi was the Chief Minister of Gujarat from 7 October 2001 to 22 May 2014, preceded by Keshubhai Patel and succeeded by Anandiben Patel.'' This can be represented by a fact where subject is ``Narendra Modi'', relation is ``position held'', object is ``Chief Minister of Gujarat'' and there are 4 qualifiers each with their qualifier relations as follows: (1) $q_1$=``7 October 2001'', $qr_1$=``start time'', (2) $q_2$=``22 May 2014'', $qr_2$=``end time'', (3) $q_3$=``Keshubhai Patel'', $qr_3$=``replaces'', and (4) $q_4$=``Anandiben Patel'', $qr_4$=``replaced by''. 

Each fact $f_i$ is encoded as a string and the overall input consists of a concatenation of such strings across all facts in $F$. The string representation for a fact $f_i$ is ``$\langle S\rangle  s_i \langle R\rangle  r_i \langle O\rangle  o_i \langle R\rangle  qr_{i_1} \langle O\rangle  q_{i_1}  \langle R\rangle  qr_{i_2} \langle O\rangle q_{i_2} \ldots$  $\langle R\rangle  qr_{i_m} \langle O\rangle  q_{i_m}$'' where $\langle S\rangle$, $\langle R\rangle$, $\langle O\rangle$ are special tokens. Finally, the overall input with $n$ facts is obtained as follows: ``generate [language] $f_1$ $f_2$ $\ldots$ $f_n$ $\langle T \rangle [t]$'' where ``[language]'' is one of our 12 languages, $\langle T \rangle$ is the section title delimiter token, and $t$ is the section title.






\noindent\underline{\textbf{Standard Transformer-based Encoder-decoder Models}}: For XF2T generation, we train multiple popular multi-lingual text generation models on Train+ Validation part of our \textsc{XAlign} dataset. We use a basic Transformer model, mT5-small finetuned on xtreme XNLI, and the IndicBART~\cite{dabre2021indicbart} for the XF2T task. We do not experiment with mBART~\cite{maurya2021zmbart} and Muril~\cite{khanuja2021muril} since their small sized model checkpoints are not publicly available. We train these models in a multi-lingual cross-lingual manner. Thus, we train a single model using training data across languages without any need for translation. 

\noindent\underline{\textbf{Bi-lingual, Multi-lingual and Translation-based models}}: Next, we experiment with different training setups. Traditionally in cross-lingual settings, it has been observed that bi-lingual models could be more accurate for some language pairs. Note that in our case, input is always in English while the output could be in any of the 12 languages. Hence, we train bi-lingual models, i.e., one model per language since our input is always in English. A drawback with this approach is the need to maintain one model per language which is cumbersome.

Further, we also train two translation based models. In the ``translate-output'' setting, we train a single English-only model which consumes English facts and generates English text. The English output is translated to desired language at test time. In the ``translate-input'' setting, English facts are translated to LR language and fed as input to train a single multi-lingual model across all languages. While translating if mapped strings for entities were present in Wikidata they were directly used. A drawback with these approaches is the need for translation at test time.

\noindent\underline{\textbf{Pretraining approaches}}: Pretraining has been a standard method to obtain very effective models even with small amounts of labeled data across several tasks in natural language processing (NLP). Domain and task specific pretraining has been shown to provide further gains~\cite{gururangan2020don}. We experiment with the following four pretraining strategies on top of the already pretrained mT5 model before finetuning it on \data{} dataset.
(1) English-only pretraining: Wang et al.~\cite{wang2021stage} provide a noisy, but larger corpus (542192 data pairs across 15 categories) crawled from Wikipedia for English F2T task. The dataset is obtained by coupling noisy English Wikipedia data with Wikidata triples. 
(2) Multi-lingual pretraining: In this method, we translate English sentences from the Wikipedia-based~\cite{wang2021stage}'s data to our LR languages. Thus, the multi-lingual pretraining data contains $\sim$6.5M data pairs. For translating sentences, we use IndicTrans~\cite{ramesh2021samanantar}. 
(3) Multi-stage pretraining: Translation is a preliminary task for effective cross-lingual NLP. Thus, in this method, in the first stage, we pretrain mT5 on translation data corresponding to English to other language pairs with $\sim$0.25M data instances per language. In the second stage, we perform multi-lingual pretraining as described above.
(4) Multi-task pretraining: This method also involves training for both translation as well as XF2T tasks. Unlike the multi-stage method where pretraining is first done for translation and then for XF2T (multi-lingual pretraining), in this method we perform the two tasks jointly in a multi-task learning setup.





\begin{figure*}
    \centering
    \includegraphics[width=\textwidth]{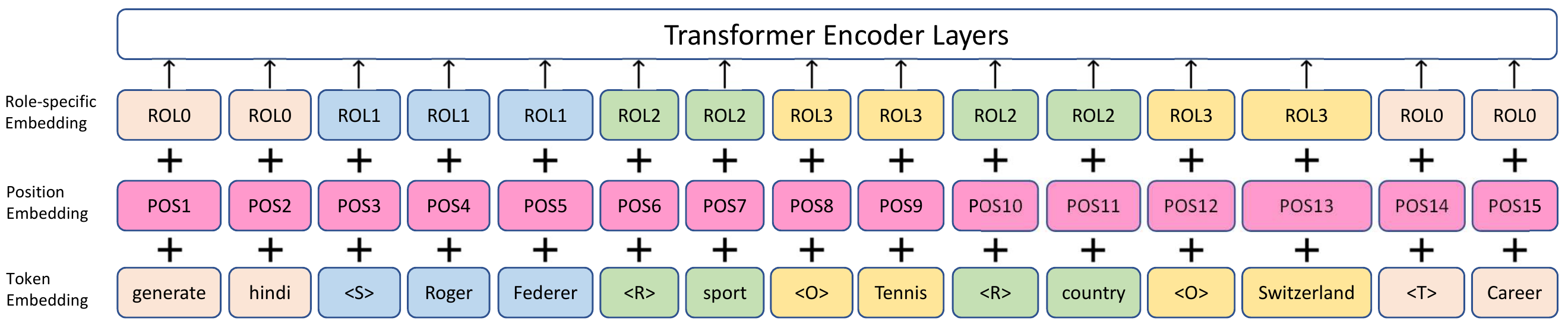}
    \caption{English facts being passed as input to mT5's encoder with token, position and (fact-aware) role embeddings.}
    \label{fig:roleEmbeddings}
\end{figure*}

\noindent\underline{\textbf{Fact-aware embeddings}}: 
The input to mT5 consists of token embeddings as well as position embeddings. For XF2T, the input is a bunch of facts. Facts contain semantically separate units each of which play a different role: subject, relation, object. We extend the standard mT5 input with specific (fact-aware) role embeddings. Specifically, we use four role IDs: 1 for subject, 2 for relation and qualifier relation, 3 for object and qualifier tokens, and 0 for everything else, as shown in Fig.~\ref{fig:roleEmbeddings}. We hope that this explicit indication of the role played by each token in the input facts, will help the model for improved XF2T generation. 

We also experimented with (1) separate role embeddings for qualifier relation and qualifier, and (2) adding fact id embeddings, i.e., if the input contains $K$ facts, we have $K$ fact IDs, and all tokens corresponding to a fact gets the same fact ID embedding. However, these did not lead to better results and thus we do not report those results.

\section{Experiments}
\label{sec:experiments}
\noindent\underline{\textbf{Implementation Details for Reproducibility}}: We closely follow Abhishek et al.~\cite{abhishek2022xalign}'s data-collection and XF2T alignment method for the creation of cross-lingual fact-to-text dataset for four additional languages. All XF2T generation approaches were run on a machine equipped with four 32GB V100 GPUs. For all experiments, we use IndicNLP~\cite{kakwani2020indicnlpsuite} to convert the low-resource languages of \data{} to the unified Devanagari script. All Transformer models have 6 encoder and 6 decoder layers. For Vanilla Transformer, we follow the standard architecture and hyper-parameters suggested in Vaswani et al.~\cite{vaswani2017attention}. For other methods,  we optimize cross entropy loss using AdamW with constant learning rate of 3e-5 with L2-norm weight decay of 0.001, batch size of 20 and dropout of 0.1. We closely follow Dabre et al.~\cite{dabre2021indicbart} for finetuning IndicBart.  

When applicable, we pretrain for 7 epochs. For English-only pretraining, we use validation subset with only English examples. For multi-lingual pretraining, we use full validation set. In multi-stage pretraining, we save best checkpoint of first stage (translation task) on validation set of translation task and use it to initialize model parameters for second stage. For multi-task pretraining, we create new validation set by combining validation set of translation task and XF2T task. We finetune for 30 epochs and use beam search with width of 4.

\noindent\underline{\textbf{Evaluation Metrics}}: We use overall BLEU scores~\cite{ramesh2021samanantar} for evaluating the multi-lingual models for English-Indic fact-sentence pairs. Following previous work, we also use METEOR~\cite{banerjee-lavie-2005-meteor} and chrF++~\cite{popovic2017chrf++}.
PARENT~\cite{dhingra2019handling} relies on the word overlap between input and the prediction text. Since the input and prediction in XF2T are in different languages, we cannot compute PARENT scores. 


\section{Results and Analysis}
\label{sec:results}

Since XF2T is a very recently proposed task, there are not many baseline methods to compare with. Abhishek et al.~\cite{abhishek2022xalign} provided very basic results using BLEU only. In this section, we will present results using methods described in Section~\ref{sec:approach}. Due to lack of space, we show per language results only for our best model. For other comparisons and analysis, we show average across all languages while pointing out any interesting per-language insights.

\noindent\underline{\textbf{Standard Transformer-based Encoder-decoder Models}}: Table~\ref{tab:stdModelResults} shows BLEU results across different (model, metric) combinations using three standard Transformer-based encoder-decoder models. Across the 12 languages, on average for each metric, mT5 performs better than IndicBART which in turn is better than vanilla Transformer. We observed that IndicBART performed exceptionally well for Bengali but is exceptionally poor on English. Given that mT5 is better on average amongst the three, we perform further experiments using mT5. 
\begin{table*}
\begin{minipage}{0.44\textwidth}
     \centering
    \scriptsize
    \begin{tabular}{|l||c|c|c|}
\hline
&BLEU&METEOR&chrF++\\
\hline
\hline
Vanilla Transformer&21.93&50.21&50.89\\
\hline
IndicBART&23.78&50.80&53.88\\
\hline
mT5&\textbf{28.13}&\textbf{53.54}&\textbf{57.27}\\
\hline
    \end{tabular}
    \caption{XF2T scores on \data{} test set using standard Transformer-based encoder-decoder models. Best results are highlighted.}
    \label{tab:stdModelResults}
\end{minipage}
\hspace{0.04\textwidth}
\begin{minipage}{0.48\textwidth}
    \centering
    \scriptsize
    \begin{tabular}{|l||c|c|c|}
\hline
&BLEU&METEOR&chrF++\\
\hline
\hline
Bi-lingual mT5 (12 models)&25.88&50.91&52.88\\
\hline
Translate-Output mT5 (1 model)&18.91&42.83&49.10\\
\hline
Translate-Input mT5 (1 model)&26.53&52.24&55.32\\
\hline
Multi-lingual mT5 (1 model)&\textbf{28.13}&\textbf{53.54}&\textbf{57.27}\\
\hline
    \end{tabular}
    \caption{XF2T scores on \data{} test set using bi-lingual, multi-lingual and translation-based variants of mT5 model. Best results are highlighted.}
    \label{tab:bi-lingualMulti-lingualTranslation}
\end{minipage}
\end{table*}

\noindent\underline{\textbf{Bi-lingual, Multi-lingual and Translation-based models}}: Table~\ref{tab:bi-lingualMulti-lingualTranslation} shows results when mT5 model is trained using various bi-lingual, multi-lingual and translation-based settings. We observe that across all settings, the initial setting of training a single multi-lingual cross-lingual model is the best on average across all metrics. That said, for Bengali, a bi-lingual model, i.e., a model specifically trained for en$\rightarrow$bn, is much better. Translate-output and translate-input settings lead to slightly improved models for English and Tamil respectively. On average, translate-output setting performs the worst while the multi-lingual setting performs the best.

\noindent\underline{\textbf{Pretraining approaches}}: Table~\ref{tab:pretraining} (lines 1 to 5) shows results using different pretraining strategies. Translation-only pretraining is the model obtained using pretraining for translation task only. We observe that multi-lingual pretraining leads to improvements compared to no XF2T specific pretraining across 3 of the 4 metrics. Multi-stage pretraining is slightly better than translation-only pretraining but not as good as multi-lingual pretraining. Finally, multi-task performs better than multi-stage. For English and Bengali, we found that multi-stage pretraining provided best results. However, multi-lingual pretraining is the best on average across languages, with biggest wins for Malayalam and Oriya. 

\begin{table*}
    \centering
    \scriptsize
    \begin{tabular}{|l|p{2.5in}||c|c|c|}
\hline
No.&Method&BLEU&METEOR&chrF++\\
\hline
\hline
1&Multi-lingual mT5 (No pretraining, no fact-aware embeddings) &28.13&53.54&57.27\\
\hline
2&Multi-stage Pretraining&27.70&51.87&55.32\\
\hline
3&Multi-task Pretraining&28.45&51.87&55.20\\
\hline
4&Translation-only  Pretraining&27.53&50.67&53.71\\
\hline
5&Multi-lingual Pretraining&28.71&\textbf{53.83}&\textbf{57.58}\\
\hline
6&Fact-aware embeddings&\textbf{29.27}&53.64&57.30\\
\hline
    \end{tabular}
    \caption{XF2T scores on \data{} test set using different pretraining strategies and fact-aware embeddings for the mT5 model. Best results  are highlighted.}
    \label{tab:pretraining}
\end{table*}

\begin{table*}
    \centering
    \scriptsize
    \begin{tabular}{|l||c|c|c||c|c|c||c|c|c|}
\hline
&\multicolumn{3}{c||}{Vanilla mT5}&\multicolumn{3}{c||}{Multi-lingual Pretraining}&\multicolumn{3}{c|}{Fact-aware embeddings}\\
\hline
&BLEU&METEOR&chrF++&BLEU&METEOR&chrF++&BLEU&METEOR&chrF++\\
\hline
hi&44.65&68.58&68.49&43.32&68.19&68.21&42.72&67.49&68.03\\
\hline
mr&26.47&56.85&59.17&27.64&56.34&57.74&29.06&55.40&57.97\\
\hline
te&14.46&43.45&52.58&15.94&42.71&52.40&16.21&42.14&51.25\\
\hline
ta&18.37&46.15&57.42&16.68&42.32&54.88&19.07&43.65&56.01\\
\hline
en&46.94&70.60&65.20&46.61&70.45&65.33&48.29&70.75&65.42\\
\hline
gu&22.69&50.31&51.36&21.39&47.98&50.14&23.27&50.00&50.64\\
\hline
bn&40.38&61.71&68.71&50.89&75.62&77.43&49.48&73.03&76.19\\
\hline
kn&10.66&32.58&46.92&11.61&33.00&47.18&11.57&33.44&46.66\\
\hline
ml&26.22&56.71&57.01&27.38&56.63&57.35&29.04&57.15&57.60\\
\hline
pa&26.96&54.82&52.33&26.04&54.17&52.50&28.65&55.19&53.38\\
\hline
or&47.17&67.82&71.20&44.97&66.49&70.64&41.75&63.77&67.96\\
\hline
as&12.61&32.93&36.91&12.00&32.04&37.15&12.16&31.61&36.44\\
\hline
Avg&28.13&53.54&57.27&28.71&53.83&57.58&29.27&53.64&57.30\\
\hline
    \end{tabular}
    \caption{XF2T scores on \data{} test set using vanilla mT5, multi-lingual pretrained mT5 and mT5 with fact-aware embedding models.}
    \label{tab:comparison}
\end{table*}

\begin{figure*}
    \centering
    \includegraphics[width=\textwidth]{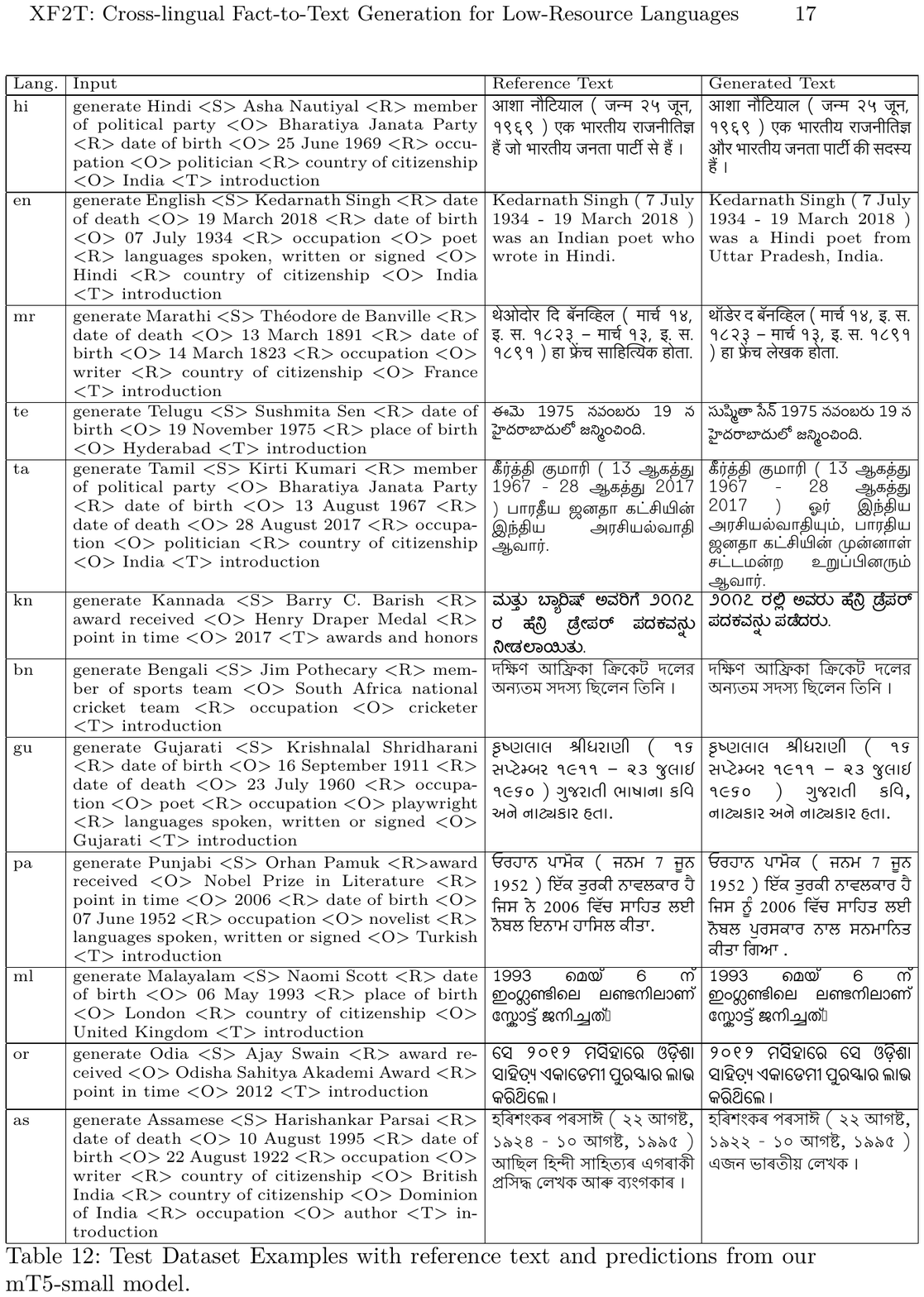}
    \captionof{table}{Test examples with reference text and predictions from our fact-aware embedding model.}
    \label{tab:caseStudies}
\end{figure*}

\noindent\underline{\textbf{Fact-aware embeddings}}: Table~\ref{tab:pretraining} (line 6) shows that fact-aware embeddings lead to improvements over the vanilla mT5 method  without fact-aware embeddings (line 1).

In summary, we note that both the proposed methods (multi-lingual pretraining, fact-aware embedding) lead to improvements over the vanilla mT5. We also experimented with combinations of these approaches but did not observe better results. Amongst these, multi-lingual pretraining performs the best on two of the metrics (METEOR and chrF++) while fact-aware embeddings perform best on BLEU. Hence, we present language-wise detailed comparison across these three models in Table~\ref{tab:comparison}. We observe that all the models perform well on bn, hi, en and or. On the other hand, performance is poor for te, ta, kn and as.

\begin{figure*}
\begin{minipage}{0.32\columnwidth}
 \centering
\includegraphics[width=\textwidth]{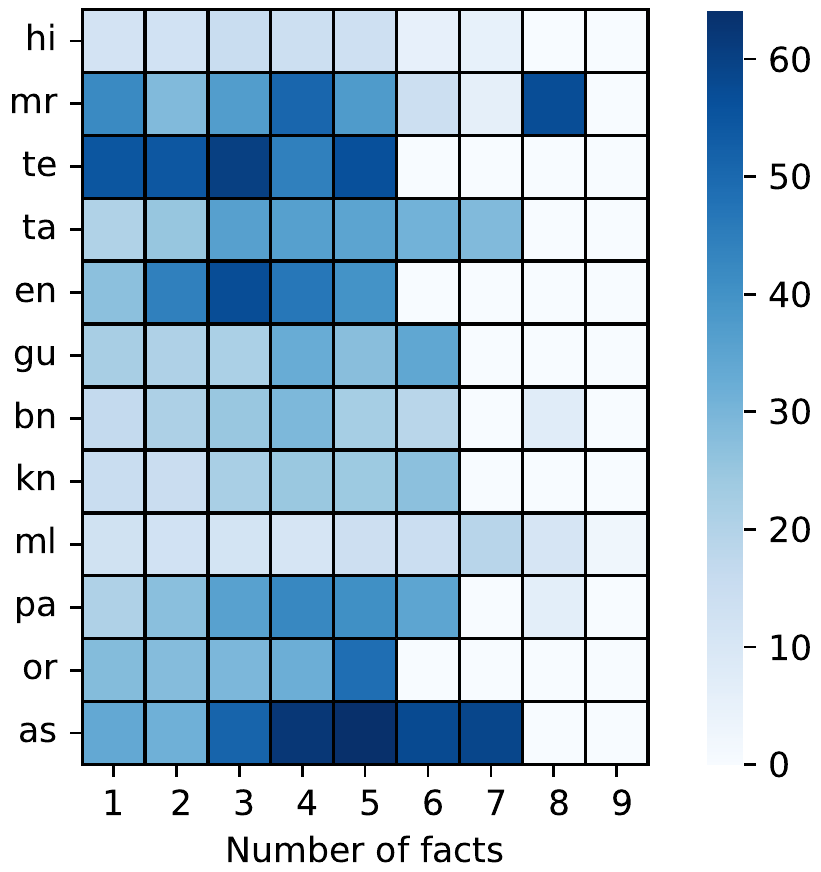}
\end{minipage}
\begin{minipage}{0.32\columnwidth}
 \centering
\includegraphics[width=\textwidth]{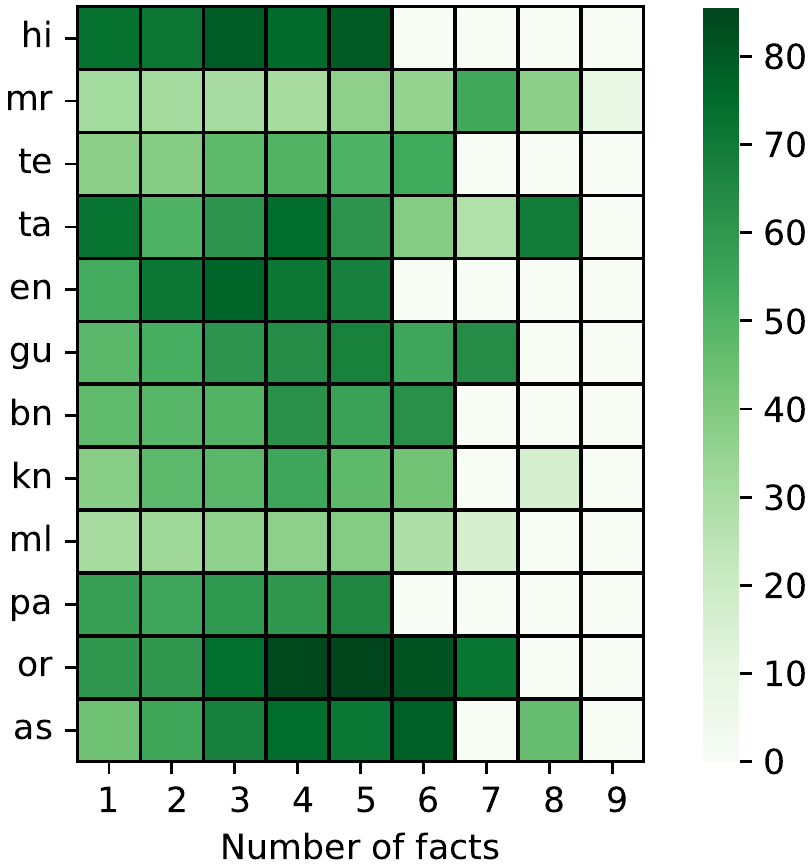}
\end{minipage}
\begin{minipage}{0.32\columnwidth}
 \centering
\includegraphics[width=\textwidth]{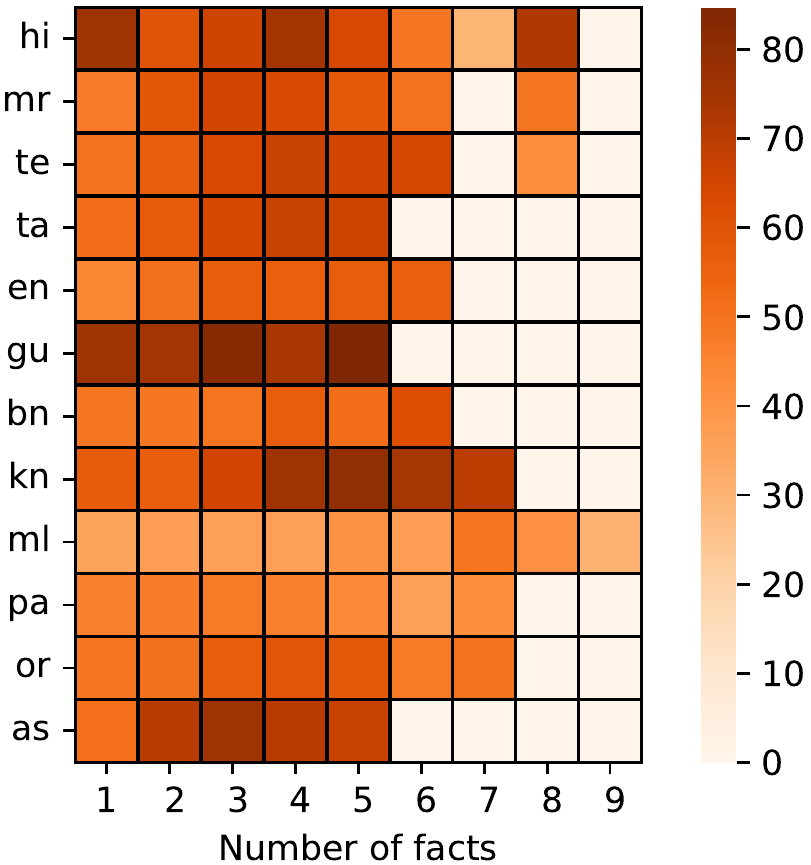}
\end{minipage}
\caption{BLEU (left), METEOR (middle) and chrF++ (right) scores for the best model across languages for test instances with a specific number of facts. White cells indicate absence of instances in that cell.}
\label{fig:factDistScores}
\end{figure*}

Fig.~\ref{fig:factDistScores} shows BLEU, METEOR and chrF++ scores for the best model across languages for test instances with a specific number of facts. Number of facts per instance range from 1 to 9. We observe that the model performs best on instances with 2--4 facts across languages and across all metrics.

Table~\ref{tab:caseStudies} shows XF2T prediction examples for our fact-aware embedding model. In general, across examples, we observe that the generated text is fluent and correct. Most of the input facts are covered by the generated sentence. Sometimes, though, the model hallucinates and brings in extra information in the output, e.g., for English, ``Uttar Pradesh'' is not mentioned as part of input facts.

\noindent\underline{\textbf{Scaling study}}: So far we presented results using small-scale models. For the fact-aware embedding model, we also train a large scale checkpoint with 12 encoder and 12 decoder layers. We observe that it leads to a BLEU of 30.90, METEOR of 55.12 and chrF++ of 59.17 which is significantly better compared to the small model as expected.

\section{Conclusion}
\label{sec:conclusion}
In this paper, we worked on the XF2T problem. We contributed the \data{} dataset which has instances with English facts aligned to 12 languages. We investigated several multi-lingual Transformer methods with different training setups, pretraining setups and input representations. We obtained models with best metrics of 30.90 BLEU, 55.12 METEOR and 59.17 chrF++ for XF2T. We make our code and dataset\footnote{\url{https://tinyurl.com/CrossF2TV2}} publicly available to empower future research in this critical area.

\if\arxiv1
\section{Ethical Concerns}
\label{sec:ethical}
We do not foresee any harmful uses of this technology. In fact, F2T generation systems are vital in many downstream Natural Language Processing (NLP) applications like automated dialog systems~\cite{wen2016multi}, domain-specific chatbots~\cite{novikova2017e2e}, open domain question answering~\cite{chen2020kgpt}, authoring sports reports~\cite{chen2008learning}, etc. We believe that these systems will be useful for powering business applications like Wikipedia text generation given English Infoboxes, automated generation of non-English product descriptions using English product attributes, etc.

As part of this work, we collected labeled data as discussed in Section~\ref{sec:dataCreation}. The dataset does not involve collection or storage of any personal identifiable information or offensive information at any stage. Human annotators were paid appropriately while performing data collection according to the standard wages set by National Translation Mission (\url{https://www.ntm.org.in/}) and mutually agreed upon. Although we have included the link to the dataset for reviewers' perusal, the data will be publicly released under MIT Open-Source License upon acceptance of this paper. The annotation exercise was approved by the Institutional Review Board of our institute.
\fi

\if\arxiv1
\section{Acknowledgements}
This research was partially funded by Ministry of
Electronics and Information Technology (MeitY),
Government of India under Sanction Order No:
11(6)/2019-HCC (TDIL). The views and conclusions contained herein are those of the authors and
should not be interpreted as necessarily representing the official policies, either expressed or implied
of MeitY.
We would like to thank the following annotators
from National Translation Mission for their crucial
contributions in creation of test dataset: Bhaswati
Bhattacharya, Aditi Sarkar, Raghunandan B. S.,
Satish M., Rashmi G.Rao, Vidyarashmi P N, Neelima Bhide, Anand Bapat, Krishna
Rao N V, Nagalakshmi DV, Aditya Bhardwaj Vuppula, Nirupama Patel, Asir. T, Sneha Gupta, Dinesh Kumar, Jasmin Gilani, Vivek R, Sivaprasad S, Pranoy J, Ashutosh Bharadwaj, Balaji Venkateshwar, Vinkesh Bansal, Ramandeep Singh, Khushi Goyal, Yashasvi LN Pasumarthy and Naren Akash.
\fi


\bibliographystyle{splncs04}
\bibliography{references}
\end{document}